\pdfoutput=1
\documentclass[10pt,twocolumn,letterpaper]{article}

\usepackage{iccv}
\usepackage{times}
\usepackage{graphicx}
\usepackage{amsmath}
\usepackage{amssymb}

\usepackage{microtype}
\usepackage{subfig}
\usepackage{booktabs} 
\usepackage{nccmath}
\usepackage{multirow}
\usepackage{enumitem}

\usepackage[font=small, labelfont=bf]{caption}

\usepackage[breaklinks=true,bookmarks=false]{hyperref}
\usepackage[accsupp]{axessibility}

\iccvfinalcopy 

\def\iccvPaperID{7} 

\ificcvfinal\fi

\newcommand{\inlineeqnum}{\refstepcounter{equation}~~\mbox{(\theequation)}}

\begin{document}

\title{perf4sight: A toolflow to model CNN training performance on Edge GPUs}

\author{Aditya Rajagopal\\
Imperial College London\\
{\tt\small adityarajagopal0@outlook.com}
\and
Christos-Savvas Bouganis\\
Imperial College London\\
{\tt\small ccb98@ic.ac.uk}
}

\maketitle
\ificcvfinal\thispagestyle{empty}\fi

\begin{abstract}
    The increased memory and processing capabilities of today's edge devices create opportunities for greater edge intelligence.
    In the domain of vision, the ability to adapt a Convolutional Neural Network's (CNN) structure and parameters to the input data distribution leads to systems with lower memory footprint,
    latency and power consumption. 
    However, due to the limited compute resources and memory budget on edge devices, it is necessary for the system to be able to predict the latency and memory footprint of the training process in order to identify favourable 
    training configurations of the network topology and device combination for efficient network adaptation.
    This work proposes perf4sight, an automated methodology for developing accurate models that predict CNN training memory footprint and latency given a target device and network. 
    This enables rapid identification of network topologies that can be retrained on the edge device with low resource consumption. 
    With PyTorch as the framework and NVIDIA Jetson TX2 as the target device, the developed models predict training memory footprint and latency with 95\% and 91\% accuracy respectively for a wide range of networks,
    opening the path towards efficient network adaptation on edge GPUs. 
\end{abstract}

\vspace{-0.2cm}
\section{Introduction}
    \label{sec:intro} 
    There is increasing demand for the optimised deployment of DNNs on edge devices. 
    Current approaches focus on optimising network weights and topology offline using datasets that represent the input data distribution that is expected to be observed at the point of deployment. 
    As it is not always possible to accurately capture the expected data distribution, the ability for a system to tune the network after deployment greatly improves its performance \cite{persephonee_2021, dapr_rajagopal2020}.
    
    Currently the adopted approach of tuning a network after deployment is to send information regarding the data distribution back to a server in order to finetune the deployed network.
    This process raises data-privacy concerns and can also be infeasible in scenarios where there is poor or no connectivity between edge device and server. 
    Consequently, there has been a push to increase the intelligence of the processing performed on edge devices \cite{edge_survey_zhou_2019}.
    For instance, the limited but increasing compute and memory capabilities of edge devices such as NVIDIA's Jetson TX2 and Xavier GPUs, have enabled the training of CNNs on edge devices.
    
    Utilising this ability, \cite{dapr_rajagopal2020} demonstrated Data-aware Pruning and Retraining (DaPR) as a viable approach to tuning the network after deployment to the data observed.
    Let us consider a scenario of a car in a city environment utilising an image classification model $\mathcal{M}_\mathcal{D}$ that has been trained on a large dataset ($\mathcal{D}$) that includes images of objects in both
    city ($\mathcal{D}'$) and country-side ($\mathcal{D}''$) environments. 
    \cite{dapr_rajagopal2020} showed that memory footprint and latency gains can be achieved by retraining a pruned version of $\mathcal{M}_\mathcal{D}$ on $\mathcal{D}'$, such that the accuracy of the 
    pruned model on the subset $\mathcal{D}'$ is at least as high as that of $\mathcal{M}_\mathcal{D}$ (unpruned) on $\mathcal{D}'$.
    Using CIFAR-100 as a proxy for $\mathcal{D}$, and various subsets of CIFAR-100 as $\mathcal{D}'$, \cite{dapr_rajagopal2020} achieve on average a 10.2pp, 2.22x and 90\% improvement in test accuracy,
    inference latency and memory footprint respectively across various networks and pruning levels.
    
    However, state-of-the-art pruning and retraining techniques such as \cite{taylor-fo_pruning_molchanov_2019, thinet_luo_2017} are far too data and compute intensive to be performed directly on an edge device.
    One possible solution to this is the Once-For-All (OFA) network proposed by \cite{ofa_2020}. 
    OFA enables quick access to a large number of high-performing pruned topologies, with significant variation in memory footprint and latency. 
    This creates the possibility for a system that can vary its network architecture over time in line with varying memory and latency requirements. 
    In a dynamic system where both the data observed and the processes running on the device can change with time, being able to select network architectures with varying resource consumptions and retraining them to
    adapt to the observed data distribution is beneficial.
    
    In order to utilise OFA in such a system, an estimate of the training memory and latency of each sampled sub-network is necessary.
    One approach to perform this estimation is to profile each architecture on the device.    
    Not only is this approach time consuming, but edge devices such as the NVIDIA Jetson TX2 tend to have shared CPU and GPU memory systems. 
    This means that a process running on one can prevent processes from starting up on the other due to a lack of available memory. 
    Such an event can be catastrophic in safety critical applications such as autonomous driving. 
    Hence there is a requirement to accurately predict the memory consumption and latency of processes without running the process itself.
    
    This work addresses the problem of accurately predicting the memory consumption and latency of training a CNN on an edge GPU to enable network adaptation on edge devices.  
    Furthermore, the applications of accurate memory and latency models extend beyond this, from Network Architecture search (NAS) to improving productivity of researchers by preventing network training deployments that fail due 
    to excessive memory consumption \cite{gao2020_dnn_mem}.
    The novel contributions of this work are: 
    \begin{enumerate}
       \item The development of performance models that utilise decision trees to predict for a given training batch size, network architecture, device and training framework - the memory consumption and latency of 
           the training process.
       
       \item A methodology to profile the training process of CNNs in order to develop models for a specific device and framework combination. 
       
       \item An extensible open-source PyTorch tool\footnote{\url{https://github.com/ICIdsl/performance_modelling.git}} (perf4sight) that automates both the profiling and modelling processes.  
    \end{enumerate}
    
    The rest of the paper is organised as follows.
    Sec.\ref{sec:background} describes the operations required to perform CNN training.
    Sec.\ref{sec:related_works} describes state-of-the-art methodologies to predict both inference and training memory and latency. 
    Sec.\ref{sec:problem_definition} introduces the modelling problem.
    Sec.\ref{sec:model_construction} describes the proposed methodology.
    Sec.\ref{sec:eval} evaluates the performance of the developed models and provides a case study that demonstrates a use case for the system. 

\section{Background}
    \label{sec:background}
    The CNN training process is based on the back-propagation algorithm that involves a forward pass to calculate the loss followed by a backward pass that calculates the gradient w.r.t. the weights and inputs per layer.
    Finally, a gradient descent step updates the weights with the calculated gradients. 
    In this process, the most memory and compute intensive operations occur in the convolution layers.
    
    %
    
    For each convolution layer the following operations are performed. For each input feature map (IFM) $x$, output feature map (OFM) $y$ and weight matrix $w$, the forward pass performs the convolution $y = x * w \inlineeqnum \label{eq:forward_pass}$\footnote{$*$ refers to the convolution operator}.
    After computation of the loss $L$, the backpropagation \cite{backprop_rumelhart_1986} algorithm requires the following computations per layer.
    The gradient w.r.t. inputs is calculated as $\frac{\delta L}{\delta x} = \frac{\delta L}{\delta y} * rot180(w) \inlineeqnum \label{eq:backward_x}$\footnote{$rot180$ refers to a $180^{\circ}$ rotation of the matrix} , and
    the gradient w.r.t. weights is calculated as $\frac{\delta L}{\delta w} = x * \frac{\delta L}{\delta y} \inlineeqnum \label{eq:backward_w}$.
    
    On NVIDIA GPUs, these operations are commonly implemented using the cuDNN \cite{cudnn_review_2019} library. There are three ways in which a convolution operation is commonly executed on a GPU.
    The \textit{Matrix Multiplication} algorithm transforms the IFM using the \textit{im2col} operation to perform a convolution as a matrix multiplication \cite{i2c_matmul_2014}.
    Alternatively, the \textit{Fast Fourier Transform} (FFT) algorithm perform convolutions as a product between weight matrices and IFMs in the frequency domain \cite{fft}. 
    It benefits from reduced operations compared to matrix multiplication and the ability to precompute and reuse the FFT of the weight matrices during the training process.
    Finally, the \textit{Winograd Convolution} algorithm reduces the number of operations performed by up to 4x compared to a matrix multiplication and uses up to 163x less temporary memory 
    compared to the FFT operation \cite{winograd_2015}.
    
    All three algorithms discussed above display variability in performance with respect to the characteristics of the layer \cite{cudnn_review_2019}. 
    This necessitates the cuDNN library to use proprietary heuristics to decide which of these three algorithms to apply on a per layer basis. 
    Consequently, models that predict the memory footprint and latency of the training process on GPUs, need to account for the memory consumption and operations of all three algorithms for each of Eq.\ref{eq:forward_pass}, \ref{eq:backward_x}, and \ref{eq:backward_w}.
        
\section{Related Works}
    \label{sec:related_works} 
        
    \subsection{Inference Performance Modelling}
        \label{sec:rel_works_inf_perf_modelling}
        Currently, the focus of performance modelling for GPUs has been on the inference stage rather than the training stage.
        State of the art approaches for inference performance modelling, \cite{augur_Lu_2019, neural_power_cai_2017}, explore the topic of predicting CNN inference memory consumption, runtime and energy on embedded GPUs. 
        \cite{augur_Lu_2019} approximate the forward pass as matrix multiplications and profile random matrix multiplication sizes on the NVIDIA TX1 device to train a model to make layer-wise predictions of memory, latency and energy.
        The developed models achieve up to 30\% prediction error in each metric across two embedded GPUs (NVIDIA TK1 and TX1) and two networks (NIN \cite{nin_Lin_2013}, VGG19M \cite{vgg_Simonyan_15}).
        
        \cite{neural_power_cai_2017} models the latency and power consumption of the inference stage of various CNNs on the NVIDIA GeForce GTX Titan X GPU (cloud system). 
        The authors develop a layer wise polynomial regression model and achieve an average error of 11.76\%, 11.66\% and 2.79\% in runtime, power and energy respectively on a wide range of networks.
        
        A layer-wise profiling system (as seen in \cite{augur_Lu_2019, neural_power_cai_2017}) obtains memory, latency and power predictions per layer and a prediction for the entire network is constructed from the layer estimates.
        This approach suits the case of modelling inference as there is only one operation performed per layer (Eq.\ref{eq:forward_pass}), allowing for the modelled attributes of a single layer to be treated in isolation.
        However in the case of training, three operations are performed at different times i.e. one on the forward pass and two on the backward pass. 
        Frameworks such as PyTorch speculatively allocate more memory than is required for the single layer even if only one layer is being profiled as it is expected that an entire network rather than a single layer will be executed when performing training.
        This makes isolating memory and latency data for a single layer during training irrelevant. 
        Alternatively, approximating all operations as matrix multiplications and profiling random sizes as in \cite{augur_Lu_2019} would lose information regarding cuDNN's heuristics that choose between the three implementations of a convolution on a layer and operation-wise (Eq.\ref{eq:forward_pass},\ref{eq:backward_x},\ref{eq:backward_w}) basis.
        As a solution, this work proposes a novel network-wise profiling strategy where each datapoint corresponds to the training of an entire network, instead of a single layer. 
        Furthermore, in contrast to \cite{augur_Lu_2019} and \cite{neural_power_cai_2017} the proposed work focuses on modelling the memory consumption and latency of the training stage instead of that of the inference stage.
    
    \subsection{Training Performance Modelling}
        \label{sec:rel_works_train_perf_modelling}
        Due to the limited investigation of efficient CNN training on edge devices, there has been little focus on modelling the latency and memory consumption of the training process on edge devices.
        On a server or distributed GPU system where training is commonly performed, the state of the art is \cite{gao2020_dnn_mem} which proposes a model to predict GPU memory consumption (but not latency) of training.
        \cite{gao2020_dnn_mem} develop an analytical model that captures both network and framework specific contributions to memory consumption during training and achieve memory prediction error rates between 0.6\% and 23\% across different frameworks (TensorFlow, PyTorch and MXNet) and networks (VGG16, ResNet50, Inception V3).  
        
        Similar to \cite{gao2020_dnn_mem}, this work uses an analytical model to capture network specific terms, i.e. weights, activations and temporary memory, required by the various convolution algorithms described in Sec.\ref{sec:background}.  
        These are static features that are unlikely to change as both the training process and methods of performing convolutions on GPUs are well established fields.
        However unlike \cite{gao2020_dnn_mem}, this work complements the analytical modelling with a profiling and learning methodology to account for the constantly changing framework and device specific terms. 
        For instance, terms that arise from the choice of cuDNN \cite{cudnn_review_2019} version are proprietary and cannot be accurately modelled through analytical approaches. 
        Furthermore, building an analytical model for targeting different frameworks (PyTorch, Tensorflow etc.) requires expert-level understanding of framework specifics and significant handcrafting of features which is both more time consuming and at risk of quickly becoming out-dated as frameworks evolve. 
        This combination of handcrafting static features and profiling for device and framework specific optimisations provides a strong argument for future generalisation over \cite{gao2020_dnn_mem}.
        
        Additionally, this work focuses on modelling training on embedded GPU systems in contrast to \cite{gao2020_dnn_mem} which focuses on server GPUs. 
        Embedded GPU systems typically have only one GPU and a unified memory system (CPU and GPU share memory) as compared to server GPU systems which can have many GPUs with each GPU having its own independent memory space. 
        Both the number of GPUs as well as the required memory management differentiate the problems of modelling embedded over server GPU systems.
        
\section{Problem Description}
    \label{sec:problem_definition}
    Motivated by the scenario presented in Sec.\ref{sec:intro}, the problem addressed by this work and the modelled attributes of the CNN training process are as follows:

    \begin{flushleft}
        Given a target device and framework, construct models that take as input a CNN network architecture and mini-batch size and accurately predict the memory consumption and latency of training.
    \end{flushleft}
    
    \vspace{-0.5cm}
    \paragraph{Training memory consumption ($\Gamma$)} The total memory consumed by the training process.
    In unified memory systems where the CPU and GPU share a memory space, this attribute also captures the memory allocated by CPU operations such as data normalisation. 
    
    \vspace{-0.3cm}
    \paragraph{Mini-batch training latency ($\Phi$)} The time taken to perform the forward and backward passes for one mini-batch of size $bs$. 
    Frameworks such as PyTorch can overlap the tasks of preparing data (eg. normalisation) for the following mini-batch with the execution of the current mini-batch. 
    Hence, only the compute time and not the time taken for data preparation is measured. 
    The time taken to perform the gradient update step in stochastic gradient descent (SGD) is also included in the measurement. 
    Total training time for a system can be estimated by multiplying $\Phi$ with the number of mini-batches of training.

\section{Model Construction}
    \label{sec:model_construction}
    The values of the attributes described in Sec.\ref{sec:problem_definition} depend on the network, training framework and target device.
    Network architecture related information can be obtained by analytically modelling the expected memory consumption and operations of training on a per layer basis. 
    However, information related to the training framework and target device such as layer wise choice of convolution algorithm (Sec.\ref{sec:background}) can only be obtained by profiling networks on the target device. 
    As such, this work proposes to use decision trees to construct high accuracy memory and latency prediction models.
    At the core of the proposed approach, analytical modelling is used to produce a set of features capturing memory footprint and computational cost of the various operations. 
    These features along with profiled system attributes (Sec.\ref{sec:problem_definition}) are used to train decision tree based prediction models.
    
    \subsection{Network-wise profiling strategy}
        \label{sec:profiling}
        \begin{figure}[t]
    \centering
    \includegraphics[width=0.5\textwidth, bb=-1 3 526 193]{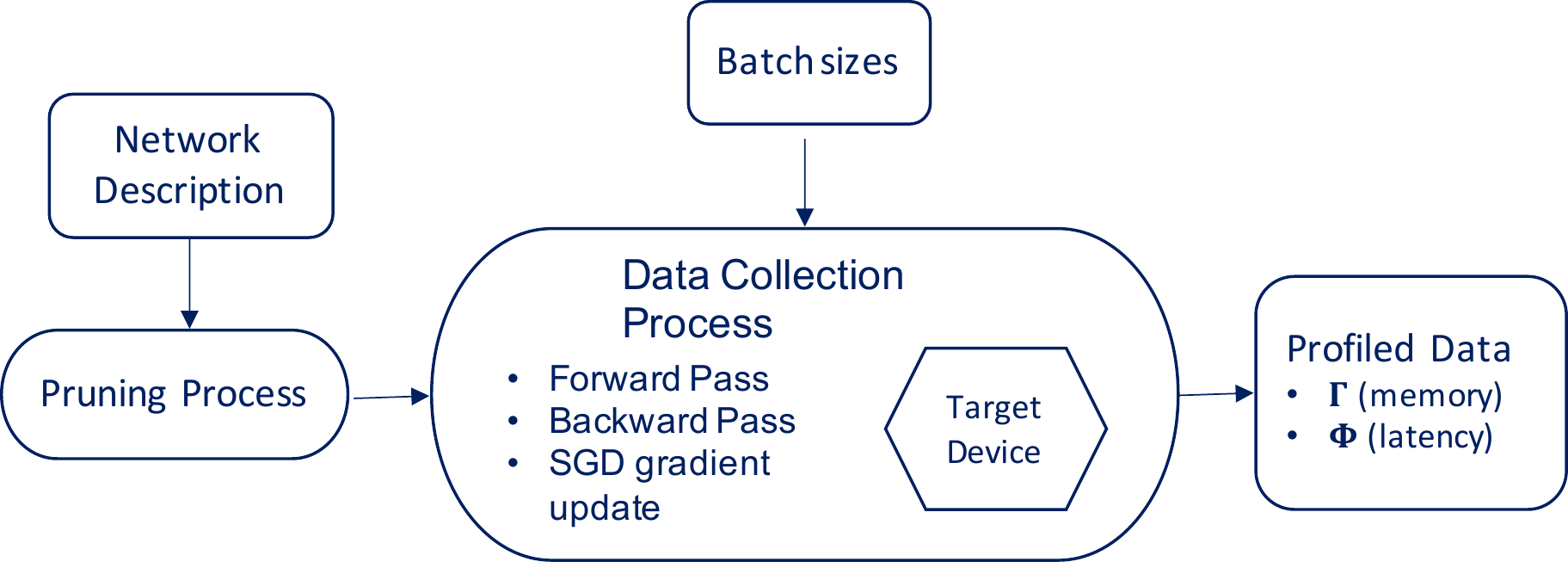}
    \caption{Components required for the proposed network-wise profiling strategy}
    \label{fig:profiling-process}
    \vspace{-0.5cm}
\end{figure}

        This section outlines the proposed methodology for profiling the desired attributes and provides details on the targeted framework and device.
        In order to encode network architecture information into the decision tree training process, structured pruning (removing entire convolution filters) was utilised as a methodology to vary the topology of a network and hence
        produce the data points used to train the decision trees.
        Fig.\ref{fig:profiling-process} shows the proposed profiling process. 
        The \textit{pruning process} takes as input the network description, the desired pruning level and strategy ($S$) and performs structured pruning.
        
        The pruned network along with $n_{bs}$ batch sizes in the range $[bs_{low},bs_{high}]$ are passed to the \textit{data collection process}.
        For each datapoint, this process profiles the Forward Pass, Backward Pass and SGD gradient update step on the target device.
        This profiling strategy limits the number of datapoints that can be collected as the degrees of freedom are the pruning levels, pruning strategies and batch size. 
        However it benefits from the fact that the information captured does not treat each layer as an isolated element but rather as part of a larger architecture where the building blocks and their 
        connectivity affects the performance. 
           
            \subsubsection{Profiling Hyperparameters}
                \label{sec:profiling_hyperparameters}  
                Sec.\ref{sec:profiling} introduced the hyperparameters of $S$, $n_{bs}$, $bs_{low}$ and $bs_{high}$, the values of which along with the targeted framework and device are detailed here. 
                The training framework targeted is PyTorch v1.6 which uses CUDA 10.2 and cuDNN 8.0.
                The target device is the NVIDIA Jetson TX2. 
                The filters to be pruned are randomly chosen ($S$), and the pruning is performed using the open-source tool ADaPT \cite{dapr_rajagopal2020}. 
                $n_{bs} = 25$, $bs_{low} = 2$ and $bs_{high} = 256$; this range and granularity of batch sizes covers the most commonly used training batch sizes (powers of 2 up till 256) while providing sufficient information in the regions inbetween. 
                Framework and device specific details on profiling can be found in Appendix \ref{app:profiling-variables}.
    
    \subsection{Decision Tree Models}
        \label{sec:perf_model}
        Four state of the art networks (ResNet18\cite{resnet_He2016}, MobileNetV2\cite{mobilenetv2_Sandler2018}, SqueezeNet\cite{squeezenet_Iandola2016} and MnasNet\cite{mnasnet_Tan2018}) were profiled for the  
        attributes defined in Sec.\ref{sec:problem_definition} using the profiling strategy described in Sec.\ref{sec:profiling} for pruning levels \{0,30,50,70,90\}\%.
        For all networks, both attributes display linearity with batch size, but varying linear fit dependent on the network architecture (pruning level). 
        Graphs of the profiled values are provided in Appendix \ref{app:model_terms}.
        
        To model this behaviour, a decision tree based approach is adopted \footnote{Linear regression was evaluated as a possibility but discarded due to poor performance}. 
        A decision tree selects terms that best partition the space into regions of low entropy. 
        Regression predictions are made by classifying new data points into these regions and predicting the mean value of that region that was learnt in the training stage \cite{dec_tree_ref}.
        This data dependent partitioning of the space inherently fits the modelling problem at hand well. 
        Hence, random forests \cite{random_forest_breiman2001} are employed to model both the memory and latency of training. 
        To do so, the modelling algorithm needs to be provided with a list of features that define the axes of the space in which the data is fit.
        These features are constructed through analytical modelling of a network architecture and are described in Sec.\ref{sec:model_terms}.
        
        \subsubsection{Analytical Modelling}
            \label{sec:model_terms}
            \begin{table}[t]
    \begin{minipage}[b]{\linewidth}
    \begin{small}
        \resizebox{\textwidth}{!}{
        \begin{tabular}{lll}
            \toprule
            \textsc{Operation} & \textsc{Feature Name} & \textsc{Feature Description} \\
            \midrule
            \multirow{4}{*}{Op Independent} & $mem_{w}$ & Weights memory \\
            & $mem_{w_{grad}}$ & Gradient w.r.t. weights memory \\
            & $mem_{ifm_{grad}}$ & Gradient w.r.t. inputs memory \\
            & $mem_{ofm_{grad}}$ & Gradient w.r.t. outputs memory \\
            \midrule
            \multirow{3}{*}{Matrix-multiplication} & $mem\_i2c^{mm}_{fwd^{total}}$ & im2col memory when storing redundant data \\
            & $mem\_i2c^{mm}_{fwd^{index}}$ & im2col memory when storing only indices \\
            ($mm$) & $ops^{mm}_{fwd}$ & Matrix multiplication operations \\
            \midrule
            \multirow{3}{*}{FFT ($fft$)} & $mem\_w_{fwd}^{fft}$ & Weights memory for FFT \\
            & $ mem\_ifm_{fwd}^{fft}$ & IFM memory for FFT \\
            & $ops_{fwd}^{fft}$ & Operations for FFT \\
            \midrule
            \multirow{2}{*}{Winograd ($wino$)} & $mem_{fwd}^{wino}$ & Memory allocated for winograd multiplication \\
            & $ops_{fwd}^{wino}$ & Operations performed for winograd multiplication \\
            \bottomrule
        \end{tabular}
    }
    \end{small}
    \caption{Description of features provided in Sec.\ref{sec:model_terms}}
    \label{tab:model_legend}
    \vspace{-0.5cm}
    \end{minipage}
\end{table}

            The models receive as input a network description and training batch size. 
            As discussed in Sec.\ref{sec:background} cuDNN uses proprietary heuristics on a per layer basis to select between the Matrix Multiplication, FFT, and Winograd convolution algorithms.
            In order to model this black-box behaviour, features corresponding to the expected memory consumption and the operations performed for all three algorithms are generated from the network description, as layer-wise algorithm choice prediction before deployment is not possible.
            For each algorithm, Eq.\ref{eq:forward_pass},\ref{eq:backward_x} and \ref{eq:backward_w} are modelled.
            
            Consider a CNN where each convolution layer $l \in \mathcal{L}$ has $n_l$ filters of size $m_l \times k_l \times k_l$. 
            Let layer $l$ have stride $s_l$, padding $p_l$ and groups $g_l$.
            Let the IFM to this layer have dimensions $bs \times m_l \times ip_l \times ip_l$, the weights $n_l \times \frac{m_l}{g_l} \times k_l \times k_l$ and the OFM $bs \times n_l \times op_l \times op_l$ where $bs$ is the batch size of training.
            The OFM spatial dimensions $op_l$ can be calculated using the equation $op_l = 1 + \lfloor \frac{ip_l + 2p_l - k_l}{s_l} \rfloor$.
            Table.\ref{tab:model_legend} describes all the features listed below.
            
            Independent of the choice of algorithm, the following features are allocated in memory.
            \begin{fleqn}
            \begin{equation*}
                \begin{split}
                    &mem_{w} = n_l \cdot \frac{m_l}{g_l} \cdot k_l^2\\ 
                    &mem_{w_{grad}} = bs \cdot n_l \cdot \frac{m_l}{g_l} \cdot k_l^2 \\
                    &mem_{ifm_{grad}} = mem_{ifm} = bs \cdot m_l \cdot ip_l^2 \\
                    &mem_{ofm_{grad}} = mem_{ofm} = bs \cdot n_l \cdot op_l^2
                \end{split}
            \end{equation*}
            \end{fleqn}
            The remaining features described below all vary with the choice of algorithm. 
            Features corresponding to the forward pass (Eq.\ref{eq:forward_pass}) will be described here.
            Those for Eq.\ref{eq:backward_x} and \ref{eq:backward_w} can be found along with a full list of features in Appendix \ref{app:model_terms}.
            \vspace{-0.3cm} 
            \paragraph{Matrix Multiplication based convolution}
                \label{sec:matmul_terms}
                A convolution is converted to a matrix multiplication by performing the \textit{im2col} operation on the IFM (LHS of the convolution).
                There are two variants of the cuDNN operator, one which stores the entire \textit{im2col} matrix in memory and one that only computes indices that allows the compute unit to read repeated values from the IFM without actually storing these duplicated values \cite{i2c_matmul_2014}.
                The proprietary nature of cuDNN means that which indices are calculated is not public. 
                However the \textit{im2col} operation reads repeated values due to the overlap between the $k_l \times k_l$ sliding windows during a convolution, where each window corresponds to 1 output pixel in $op_l$.
                Within a sliding window, the offsets of the $k_l^2$ values from the top-left element of the window ($win[0]$) are fixed regardless of which window ($op_l$) is being computed. 
                However, $s_l$ and $p_l$ decide the pattern of $win[0]$.
                A reasonable assumption is that the indices computed to reduce the memory overhead of the \textit{im2col} operation are the location of all $win[0]$ for each image in the IFM. 
                The operations do not change between these two variants.
                Under these assumptions, the parameter count and operations for Eq.\ref{eq:forward_pass} are: 
                
                \begin{fleqn}
                \begin{equation*}
                    \begin{split}
                        &mem\_i2c_{fwd^{total}}^{mm} = bs \cdot op_l^2 \cdot k_l^2 \cdot m_l \\
                        &mem\_i2c_{fwd^{index}}^{mm} = bs \cdot op_l^2 \\
                        &ops_{fwd}^{mm} = bs \cdot n_l \cdot op_l^2 \cdot k_l^2 \cdot \frac{m_l}{g_l} \\
                    \end{split}
                \end{equation*}
                \end{fleqn} 
            
            \vspace{-0.5cm} 
            \paragraph{FFT based convolution} 
                \label{sec:fft_terms}
                As provided in \cite{fft}, the features required to model Eq.\ref{eq:forward_pass} are: 
                \begin{fleqn}
                \begin{equation*}
                    \begin{split}
                        & mem\_w_{fwd}^{fft} = n_l \cdot \frac{m_l}{g_l} \cdot ip_l \cdot (1 + ip_l) \\
                        & mem\_ifm_{fwd}^{fft} = bs \cdot m_l \cdot ip_l \cdot (1+ip_l) 
                    \end{split}
                \end{equation*}
                \vspace{-0.25cm} 
                \begin{alignat*}{2}
                    & ops_{fwd}^{fft} = ip_l^2 \cdot log(ip_l) \cdot & (bs \cdot (m_l + n_l) + n_l \cdot \frac{m_l}{g_l}) \\ && + bs \cdot n_l \cdot m_l \cdot ip_l^2 
                \end{alignat*}
                \end{fleqn}
                 
            \vspace{-0.5cm} 
            \paragraph{Winograd based convolution}
                \label{sec:winograd_terms}
                The Winograd minimal filtering algorithm \cite{winograd1980arithmetic} reduces the number of operations required to calculate $q$ outputs using an $r$-tap FIR filter. 
                As provided in \cite{winograd_2015}, for the computation of $q \times q$ outputs with an $r \times r$ convolutional filter, the 2-D Winograd minimal filtering algorithm is 
                $Y = A^T \Big[[GgG^T] \odot [B^TdB] \Big] A \inlineeqnum \label{eq:winograd}$.
                
                In Eq.\ref{eq:winograd}, $Y$ has shape $(q \times q)$, $g$ is an $(r \times r)$ convolutional filter, and $d$ is a tile of the input matrix of size $(q+r-1 \times q+r-1)$. 
                $A$, $G$ and $B$ are transform matrices that only depend on the values of $q$ and $r$. 
                The algorithm takes $(q+r-1)^2$ multiplications to produce $q \times q$ output values.
                
                In Eq.\ref{eq:forward_pass}, each channel of the IFM $x$ is split into $\lceil \frac{ip_l}{q} \rceil^2$ tiles and each channel of the filter $w$ is split into $\lceil \frac{k}{r} \rceil^2$ tiles.
                Each IFM and filter tile corresponds to $d$ and $g$ respectively in Eq.\ref{eq:winograd}.
                The memory consumed by $A$, $G$ and $B$ is fixed and does not scale with any of the layer parameters. 
                The Hadamard product contains $3 \cdot (q+r-1)^2$ parameters (one for the LHS, RHS and result).
                The memory and operations scale with layer parameters depending on the number of these products that can be done in parallel.
                Each of the $\lceil \frac{ip_l}{q} \rceil^2$ tiles, $n_l$ filters and $bs$ images can be computed in parallel as they constitute independent operations. 
                Accumulation needs to occur across $\lceil \frac{k}{r} \rceil^2$ filter tiles and $m_l$ IFM channels and hence do not lend themselves to parallelisation. 
                Eq.\ref{eq:winograd} computes $(q+r-1)^2$ multiplications, hence the parameter and operations counts for Eq.\ref{eq:forward_pass} are: 
                
                \begin{fleqn}
                \begin{equation*}
                    \begin{split}
                        & mem_{fwd}^{wino} = bs \cdot n_l \cdot \lceil \frac{ip_l}{q} \rceil^2 \cdot 3 \cdot (q+r-1)^2 \\
                        & ops_{fwd}^{wino} = bs \cdot n_l \cdot \frac{m_l}{g_l} \cdot \lceil \frac{ip_l}{q} \rceil^2 \cdot \lceil \frac{k}{r} \rceil^2 \cdot (q+r-1)^2 
                    \end{split}
                \end{equation*}
                \end{fleqn}
                

    \subsection{Training Memory and Latency Models}
        \label{sec:perf_model_mem_lat}
        \begin{figure}[t!]
    \centering
    \includegraphics[trim={0 0 0 0},clip, width=0.4\textwidth, bb=-1 1 353 170]{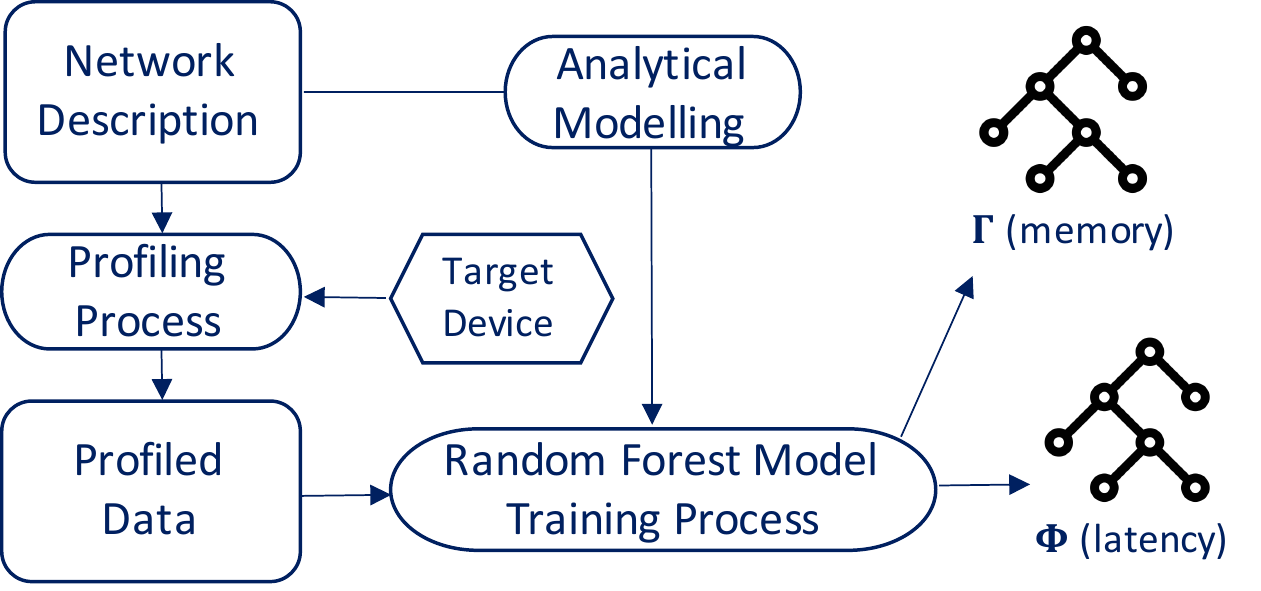}
    \caption{High-level description of the entire perf4sight toolflow.}
    \label{fig:toolflow}
    \vspace{-0.5cm}
\end{figure}

        For a given batch size, network architecture and device, the $mem$ and $ops$ features detailed in Appendix \ref{app:model_terms} are calculated per-layer and summed across all layers to 
        obtain an estimate for the network.
        These set of 42 features are provided to the random forest trainer along with training data consisting of profiled $\Gamma$ and $\Phi$ values for a range of batch-sizes and pruning levels 
        (detailed in Sec.\ref{sec:profiling_hyperparameters}).
        A separate random forest model is developed for each of the target attributes.
        This process is shown in Fig.\ref{fig:toolflow}.

\section{Evaluation}
    \label{sec:eval}
    This section describes the construction of the training and test sets for the evaluation and evaluates the performance models in Fig.\ref{fig:toolflow} on the following criteria. 
    First the models are evaluated on their ability to predict the two attributes when the training and test sets have the same networks, but different topologies.
    The topologies are varied between the training and test sets by having different pruning levels in each.
    The evaluation is then extended to allow the network itself to change, exploring the idea of training performance prediction models on data from a "basis" of networks and predicting attribute values for networks not 
    in the basis. 
    The final section presents a case study describing the model selection and retraining use case presented in Sec.\ref{sec:intro}.
    
    \subsection{Constructing training and test sets}
        \label{sec:eval_hyperparams}
        The degrees of freedom when generating configurations to profile are pruning levels, pruning strategies and batch-size. 
        With batch-sizes and pruning strategy fixed to those specified in Sec.\ref{sec:profiling_hyperparameters}, the training and test sets are developed by varying the pruning levels in each. 
        Let the set of all pruning levels considered be $\{5x\ | x \in [0,18]\}$.
        The training set is constructed by finding the smallest set of pruning levels that captures sufficient information regarding the network topology such that it can predict the attribute values of other topologies of the same
        network well.
        
        AlexNet \cite{alexnet} was used to tune the training set size hyperparameter. 
        Sizes of training set ($T$) from 1 ($T = \{0\}$) to 8 ($T = \{0,10,20,30,50,60,70,90\}$) in increments of 1 were used to train the two models and the resulting accuracy evaluated on a 
        test set $\{5x | x \in [0,18], 5x \notin T\}$. 
        For $T = \{0\}$, the test error varied between 33\%-74\% across the attributes and decreased until $T = \{0,30,50,70,90\}$ after which it plateaued at 3\%-6\%. 
        Expecting this trend to hold for other networks, this is the training set chosen for the following experiments, with the test set containing pruning levels $\{5x | x \in [0,18], 5x \notin T\}$.
        To avoid biasing the results, AlexNet is not used in the rest of the evaluation.
        
    \subsection{Same base network in training and test sets}
        \label{sec:eval_seen_net_seen_pl}
        \begin{figure}[t]
    \begin{minipage}[b]{\linewidth}
    \includegraphics[width = \textwidth, bb=0 0 406 224]{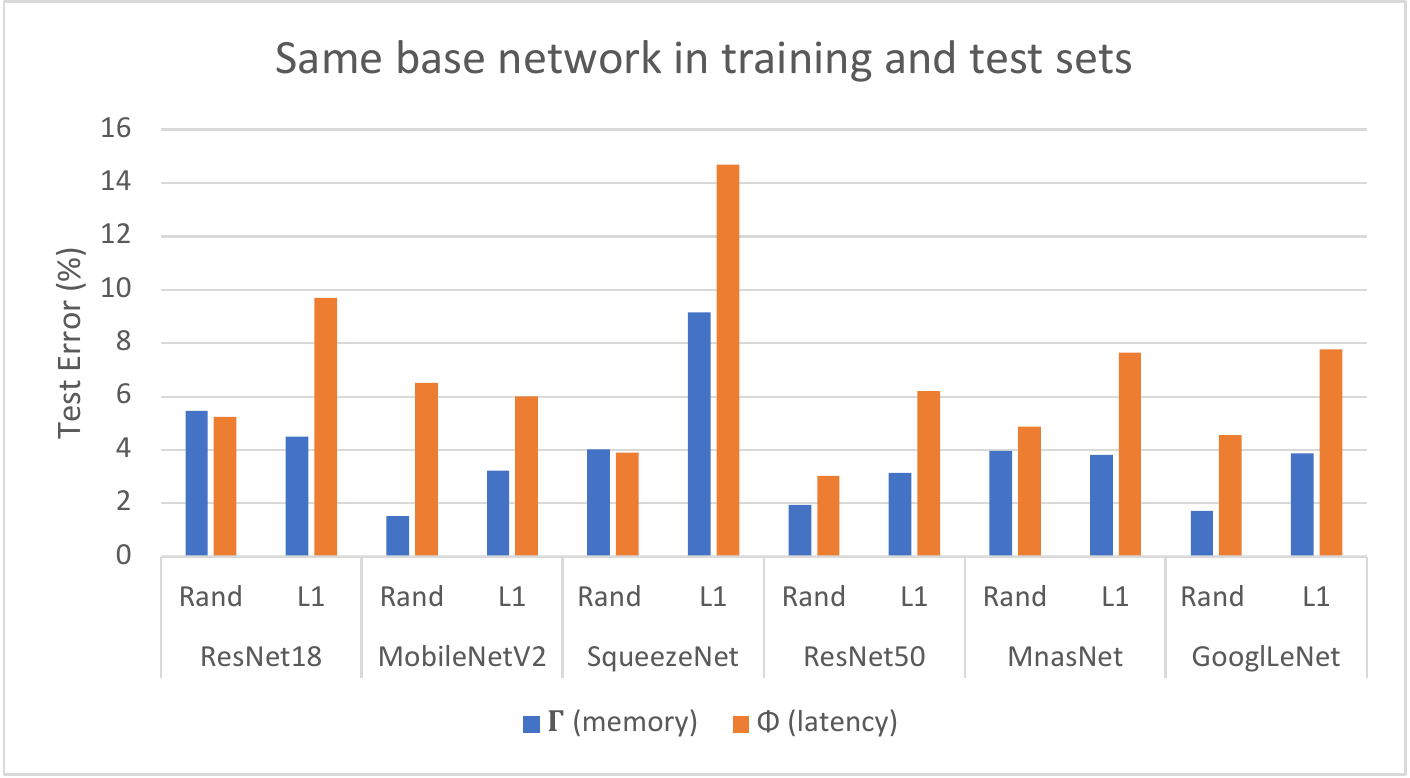} 
    \caption{Mean test error of attribute prediction for random and L1-norm pruning strategies across 25 batch sizes in the range [2,256] and pruning levels \{$5x | x \in [0,18], 5x \notin \{0,30,50,70,90\}$\}}
    \label{fig:test_errors}
    \vspace{-0.3cm}
    \end{minipage}
\end{figure}

        All results presented in this section correspond to the scenario where both the training and test sets are constructed from data from the same base network but differ in topology (pruning levels and strategy).
        Random pruning strategy refers to randomly pruning filters with equal probability across all layers. 
        
        Bars labelled \textit{Rand} in Fig.\ref{fig:test_errors} show the mean attribute prediction error averaged across a wide range of pruning levels and batch-sizes when a random pruning strategy was employed for both 
        training and test sets. 
        Bars labelled \textit{L1} in Fig.\ref{fig:test_errors} show the same when a random pruning strategy was employed for the training set, 
        but an L1-norm pruning strategy for the test set. 
        This strategy prunes filters with the smallest L1-norm first and results in more filters pruned from deeper layers.
        With a comprehensive coverage including both human-designed and NAS generated networks, the results show that the mean prediction error of $\Gamma$ and $\Phi$ does not exceed 9.15\% and 14.7\% respectively with only 
        small increases in error in most cases when testing on L1-norm pruned networks. 
        
        %
        %
        To further explore the ability of the modelling to encode network topology information, MobileNetV2 was pruned to 50\% with 100 random pruning strategies including uniform pruning across all layers 
        and increased pruning at early, late or middle layers.
        For batch-size 80, the mean and variation in attribute values $\Gamma$ and $\Phi$ across topologies were 4423$\pm$1597 MB and 1741$\pm$871 ms respectively. 
        The models, trained on just a uniform random pruning strategy, predicted these attributes with a mean error across topologies of 1.32\% and 9.90\% respectively, 
        which are comparable to those seen in Fig.\ref{fig:test_errors}.
        The results from this section demonstrate the ability of the proposed methodology to capture network architecture dependent information well. 
        
        \subsubsection{State-of-the-art comparison}
            \label{sec:eval_sota_comparison}
            Compared to works focusing on inference stage prediction (\cite{neural_power_cai_2017, augur_Lu_2019}), which achieve error rates between 12-30\%, the results displayed here outperform these methods on the more complex problem of training. 
            
            DNNMem \cite{gao2020_dnn_mem} predicts the memory consumption of training for an NVIDIA Tesla P40 GPU using PyTorch 1.2, CUDA 9.0 and CuDNN 7.0.2. 
            Due to lack of access to a P40 or DNNMem's source code, but to perform as fair a comparison as possible, ResNet50 was profiled on an NVIDIA RTX 2080Ti server GPU using PyTorch 1.6, CUDA 10.2 and CuDNN 7.6 as the framework configurations used in \cite{gao2020_dnn_mem} do not support the 2080Ti. 
            $\Gamma$ (corresponding to DNNMem's memory consumption attribute) had a 2.45\% error across batch-sizes and topologies (pruning levels). 
            This significantly outperforms \cite{gao2020_dnn_mem} which had a 17.4\% error across a range of batch-sizes and input image sizes.
            Possible reasons for the worse error achieved by \cite{gao2020_dnn_mem} could be due to either inaccuracies in handcrafting features per training framework or the fact that \cite{gao2020_dnn_mem}'s 
            model accounts for a multiple GPU setting which may have lead to inaccuracies on the single GPU setting evaluated here. 
            By demonstrating perf4sight's ability to generalise from a unified memory embedded GPU system to a non-unified memory single GPU server system, these results indicate potential for future generalisation to other
            devices and training frameworks.
        
    \subsection{Different base networks in training and test sets}
        \begin{figure}[t]
    \begin{minipage}[b]{\linewidth}
    \includegraphics[width = \textwidth, bb=0 0 409 224]{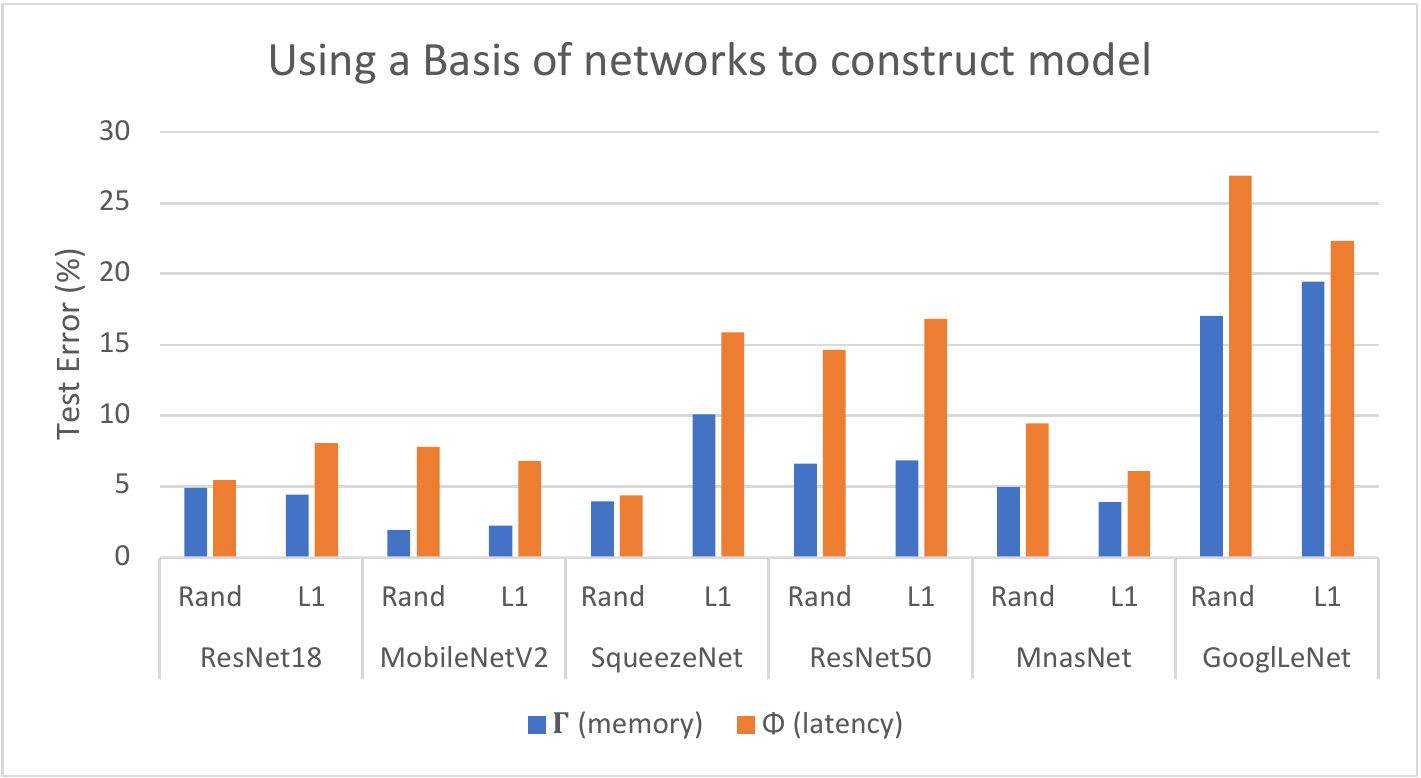} 
    \caption{Mean test error of attribute prediction across 25 batch sizes in the range [2,256] and pruning levels \{$5x | x \in [0,18]$\}. The random forest models used here are trained using data from a basis of networks containing ResNet18,
             MobileNetV2, and SqueezeNet.}
    \label{fig:basis_test_errors}
    \vspace{-0.5cm}
    \end{minipage}
\end{figure}

        \label{sec:eval_unseen_net_unseen_pl}
        This section investigates the performance of the proposed framework in the case where the targeted network is not known beforehand but rather a model is built using data from a "basis" of networks, 
        and predictions are made for the unseen network.
        The training set was constructed with a combination of data from ResNet18, MobileNetV2 and SqueezeNet (basis) for the same batch-sizes and pruning levels listed in Sec.\ref{sec:eval_hyperparams} and a 
        uniform random pruning strategy.
        Testing was performed on data from both random and L1-norm pruned networks.
        The attribute prediction errors are shown in Fig.\ref{fig:basis_test_errors}.
        
        Across all attributes and pruning strategies, networks present in the basis (ResNet18, MobilNetV2 and SqueezeNet) showed -1pp\footnote{pp stands for percentage points}, +4.6pp, +2.7pp mean increase in error respectively 
        and networks not present in the basis (ResNet50, MnasNet and GoogLeNet \cite{googlenet_Szegedy2015}) showed +5.6pp, +2.55, +16pp mean increase in error respectively compared to the results in Fig.\ref{fig:test_errors}. 
        The results support the idea of using a basis of networks, but the significant error degradation of GoogLeNet suggest that best performance is achieved when the exact building blocks if not the 
        identical network is present in both training and test sets. \footnote{Across Fig.\ref{fig:test_errors},\ref{fig:basis_test_errors}, the mean errors for 
        $\Gamma$ and $\Phi$ are 5.53\% and 9.37\%.}
        Appendix \ref{app:basis_investigation} provides a breakdown of the building blocks of all the networks.

    \subsection{Case Study: On-device OFA}
        \label{sec:case_study}
        \begin{table*}[t]
    \begin{minipage}[b]{\linewidth}
    \begin{small}
    \resizebox{\textwidth}{!}{
        \begin{tabular}{ccllllcccccccc}
            \toprule
            \multicolumn{14}{c}{\textbf{\textsc{Performance gains from on-device model selection and retraining}}} \\
            \cmidrule(lr){1-14}
             & & & & & & \multicolumn{8}{c}{\textbf{Top1 Test Accuracy on subset (\%)}} \\
            \cmidrule(lr){7-14}
            \multirow{2}{*}{\textbf{Sub-network}} & \multicolumn{1}{c}{\multirow{2}{*}{Search Time}} & \multirow{2}{*}{Model Size} & \multirow{2}{*}{$\Gamma$} & \multirow{2}{*}{$\gamma$} & \multirow{2}{*}{$\phi$}\
                                                    & \multicolumn{2}{c}{City} & \multicolumn{2}{c}{Off-road} & \multicolumn{2}{c}{Motorway} & \multicolumn{2}{c}{Country-side} \\
            \cmidrule(lr){7-14}
            & \multicolumn{1}{c}{(hours)} & (MB) & (MB) & (MB) & (ms) & Initial & Retrained & Initial & Retrained & Initial & Retrained & Initial & Retrained \\
            \cmidrule(lr){1-14}
            MAX & - & 192 (1x) & 5838 (1x) & 1958 (1x) & 69.6 (1x) & 82.0 & - & 86.2 & - & 78.3 & - & 82.4 & - \\
            \cmidrule(lr){1-14}
            A & 382/1.9 & 153 (1.3x) & 3735 (1.6x) & 1873 (1.05x)& 39.2 (1.8x) & 81.4 & 82.8 (+0.8pp) & 83.9 & 90.4 (+4.2pp) & 78.0 & 81.0 (+2.7pp) & 81.7 & 83.6 (+1.2pp) \\
            \cmidrule(lr){1-14}
            B & 519/2.6 & 76 (2.5x) & 3104 (1.9x) & 1733 (1.1x) & 25.1 (2.8x) & 79.6 & 81.6 (-0.4pp) & 84.1 & 89.9 (+3.7pp) & 76.4 & 80.2 (+1.9pp) & 80.0 & 81.9 (-0.5pp) \\
            \cmidrule(lr){1-14}
            MIN & - & 26 (7.4x) & 2768 (2.1x) & 1569 (1.2x) & 19.1 (3.6x) & 76.4 & 78.9 (-3.1pp) & 79.6 & 88.1 (+1.9pp) & 70.8 & 77.3 (-1.0pp) & 77.0 & 79.4 (-3pp) \\
            \bottomrule
        \end{tabular}
    }
    \end{small}
    \caption{Top1 test accuracy, model search time, model size, $\Gamma$, $\gamma$ and $\phi$ for various sub networks of OFAResNet50 on the 4 autonomous driving subsets.
             The search time column displays the results of the (naive / modelling) approaches. 
             The comparisons presented in brackets are all relative to model MAX.
             Initial and Retrained refer to the Top1 test accuracy on the subset with and without retraining the searched model for 1 epoch on that subset respectively.
             $\Gamma$ is reported for batch size 32 and $\gamma$ and $\phi$ reported are for batch size 1.}
    \label{tab:case_study}
    \vspace{-0.5cm}
    \end{minipage}
\end{table*}

        The Once-for-All (OFA) network \cite{ofa_2020} is a large super network trained once, following which a large number of smaller sub-networks can be sampled, each with good performance on the original training dataset.
        This case study explores the option of using the OFA network on an edge device as a way to perform on-device NAS in order to accommodate a dynamic system where both the observed data distribution and available memory and
        latency budgets can vary over time.
        The OFA network used is OFAResNet50 trained and open-sourced by \cite{ofa_2020} and has the same building blocks as ResNet50, but a slightly different connectivity. 

        \vspace{-0.4cm}
        \paragraph{Dynamic system description} Let us consider the scenario of an autonomous car performing image classification on an NVIDIA Jetson TX2.
        As examples of input data for classification, let us consider 4 subsets of the ILSVRC'12 dataset containing images that could be observed by an autonomous car in city, motorway, country-side and off-road settings.
        Details of the subsets are provided in Appendix \ref{app:subset_details}.
        Let us consider the case where the observed data distribution changes over time from a city to the country-side and the memory consumption of other, possibly safety critical, processes running on the device 
        also change over time. 
        Assume there is some available time during which the system can retrain a deployed network to account for the shift in the observed data distribution.
        Let us place hard constraints on the following three attributes: acceptable memory consumption of training ($\Gamma$) and inference ($\gamma$), and latency ($\phi$) of inference, all of which can change over time.

        Given the above scenario, the goal of this case study is to describe a system that can safely and quickly identify network architectures on which training and inference can be performed without exceeding 
        the above constraints. 
        The proposed system uses the evolutionary search (ES) algorithm proposed in \cite{ofa_2020} to search the OFA super network for a sub-network that meets the requirements.
        The ES algorithm starts with a population of 100 sub-networks and runs 500 iterations of evolutionary search before identifying the best performing sub-network within the constraints.
        This results in the process sampling at least $50,000$ sub-networks, but often more due to some sub-networks exceeding the hard constraints.
        Each sub-network sampled requires an estimation of the three constrained attributes. 
        The following analysis compares both the approach of profiling each sub-network on device and utilising the developed memory models.

        \vspace{-0.4cm}
        \paragraph{Naive Approach (Profiling)} It is not feasible to profile offline the attribute values for all possible sub-networks as the weight sharing employed by OFA results in too many possible sub-networks to do so. 
        Furthermore, the mean and standard deviation of $\Gamma$ for a 100 sub-networks sampled from OFAResNet50 profiled on the TX2 for batch sizes 32, 64 and 128 was 4318 $\pm$ 1129 MB.
        This large variation means that a single estimate is also not feasible for all possible sub-networks sampled, thus necessitating a case-by-case sampling on-device.
        In the case where there are other safety critical applications to be executed, profiling is not feasible due to the possibility of a profiled sub-network consuming more memory than 
        is acceptable and preventing the start up of another application.
        Additionally, reliably profiling requires use of the GPU and multiple runs to average across.
        On the TX2 this process takes on average 20s per data point.
        With the $50,000$ sub-networks sampled by the ES algorithm, this would result in an infeasible 11 days of run-time. 

        \vspace{-0.4cm}
        \paragraph{Utilising performance models} Alternatively, utilising the memory prediction model to predict the attribute values only requires 0.1s and 2MB as it simply requires the inference of a random forest model. 
        For the same search space as above, this would result in 1.4 hours of run-time with the entire process running only on the CPU. 
        The model for $\Gamma$ developed in Sec.\ref{sec:eval_seen_net_seen_pl} using data from ResNet50 generalises well to the OFA version and predicts $\Gamma$ for the 100 sampled sub-networks 
        with a mean error of 4.28\%. 
        
        Furthermore, as the problem of modelling inference stage attributes is a subset of that of training, random forest models were developed to predict both $\gamma$ and $\phi$.
        Batched (throughput driven) is less common than single image (latency driven) inference on the edge device setting as there are diminishing throughput gains after batch sizes greater than 32 on the
        Jetson TX2 \cite{benchmarking_ai_apps_2021}.
        Thus profiling for the batch sizes specified in Sec.\ref{sec:profiling_hyperparameters} yields little useful information.
        Limiting the batch sizes profiled to between 1 and 32 and utilising only the five pruning levels specified in Sec.\ref{sec:eval_hyperparams} results in too few data points to train models on.
        Instead, the models developed for $\gamma$ and $\phi$ are trained using profiled data for 25 out of the 100 sampled OFA sub-networks for batch sizes 1,2,4,8,16,32 and the reported test errors are for the remaining 75
        sampled networks averaged across all batch sizes.
        The profiling was performed only for the inference stage and only the features corresponding to the forward pass from Appendix \ref{app:model_terms} were used to train the random forest models. 
        The developed models predict $\gamma$ and $\phi$ with 1.8\% and 4.4\% errors respectively.

        \vspace{-0.4cm}
        \paragraph{Performance gains} 
        Table \ref{tab:case_study} displays the results of retraining 4 different sub-networks on the 4 subsets described above.
        The sub-networks MAX and MIN correspond to the largest and smallest sub-networks that can be extracted from OFAResNet50 and hence no search was required to obtain them.
        Sub-networks A and B are obtained using the ES algorithm with progressively stricter constraints on $\Gamma$, $\gamma$ and $\phi$.
        The MAX sub-network is used as a comparison as this is assumed to be the best possible model that can be deployed when maximising for accuracy without retraining.
        
        The results show that not performing evolutionary search (MIN) produces models that consistently under perform compared to MAX.  
        Performing evolutionary search (A and B) results in model size, $\Gamma$, $\gamma$, $\phi$ improvements up to 2.5x, 1.9x, 1.1x and 2.8x respectively with better Top1 test 
        accuracy, after retraining, compared to MAX in most cases.
        Furthermore, performing evolutionary search using the naive approach of profiling results in infeasible search times for on-device deployment while utilising the models provides a 200x improvement in search time allowing
        the process of model selection and retraining to be performed on-device.
        
\section{Conclusion}
    \label{sec:conc}
    This work proposes perf4sight, a methodology to model the memory consumption and latency of the training process for a given combination of network, device and framework. 
    The developed decision tree based models predict memory consumption and latency of the training process with a mean error of 5.53\% and 9.37\% respectively over a wide range of 
    network topologies and mini-batch sizes for the Jetson TX2 device and PyTorch framework.
    These error rates are a significant improvement over those obtained by other works when modelling the same attributes on both CNN training and the simpler problem of inference. 
    Thus, perf4sight enables the quick and accurate identification of favourable training configurations on edge devices.
    The methodology has been automated and open-sourced to enable researchers working on topics from NAS to resource constrained training of CNNs to benefit from the ability to accurately predict hardware attributes of CNN
    training.

{\small
\bibliographystyle{ieee_fullname}
\bibliography{main.bib}

\begin{thebibliography}{10}\itemsep=-1pt

\bibitem{random_forest_breiman2001}
Leo Breiman.
\newblock Random forests.
\newblock {\em Machine Learning}, 45(1):5--32, 2001.

\bibitem{neural_power_cai_2017}
Ermao Cai, Da-Cheng Juan, Dimitrios Stamoulis, and Diana Marculescu.
\newblock Neuralpower: Predict and deploy energy-efficient convolutional neural
  networks.
\newblock {\em arXiv preprint arXiv:1710.05420}, 2017.

\bibitem{ofa_2020}
Han Cai, Chuang Gan, Tianzhe Wang, Zhekai Zhang, and Song Han.
\newblock Once for all: Train one network and specialize it for efficient
  deployment.
\newblock In {\em International Conference on Learning Representations}, 2020.

\bibitem{i2c_matmul_2014}
Sharan Chetlur, Cliff Woolley, Philippe Vandermersch, Jonathan Cohen, John
  Tran, Bryan Catanzaro, and Evan Shelhamer.
\newblock cudnn: Efficient primitives for deep learning, 2014.

\bibitem{gao2020_dnn_mem}
Yanjie Gao, Yu Liu, Hongyu Zhang, Zhengxian Li, Yonghao Zhu, Haoxiang Lin, and
  Mao Yang.
\newblock Estimating gpu memory consumption of deep learning models.
\newblock In {\em Proceedings of the 28th ACM Joint Meeting on European
  Software Engineering Conference and Symposium on the Foundations of Software
  Engineering}, ESEC/FSE 2020, page 1342–1352, New York, NY, USA, 2020.
  Association for Computing Machinery.

\bibitem{resnet_He2016}
Kaiming He, Xiangyu Zhang, Shaoqing Ren, and Jian Sun.
\newblock Deep residual learning for image recognition.
\newblock volume 2016-December, pages 770--778. IEEE Computer Society, 12 2016.

\bibitem{squeezenet_Iandola2016}
Forrest~N. Iandola, M. Moskewicz, Khalid Ashraf, Song Han, W. Dally, and K.
  Keutzer.
\newblock Squeezenet: Alexnet-level accuracy with 50x fewer parameters and
  $<$1mb model size.
\newblock {\em ArXiv}, abs/1602.07360, 2016.

\bibitem{cudnn_review_2019}
Marc Jorda, Pedro Valero-Lara, and Antonio~J. Pena.
\newblock Performance evaluation of cudnn convolution algorithms on nvidia
  volta gpus.
\newblock {\em IEEE Access}, 7:70461--70473, 2019.

\bibitem{benchmarking_ai_apps_2021}
Pilsung KANG and Jongmin JO.
\newblock Benchmarking modern edge devices for ai applications.
\newblock {\em IEICE Transactions on Information and Systems}, E104.D:394--403,
  03 2021.

\bibitem{alexnet}
Alex Krizhevsky, Ilya Sutskever, and Geoffrey~E Hinton.
\newblock Imagenet classification with deep convolutional neural networks.
\newblock In F. Pereira, C.~J.~C. Burges, L. Bottou, and K.~Q. Weinberger,
  editors, {\em Advances in Neural Information Processing Systems}, volume~25.
  Curran Associates, Inc., 2012.

\bibitem{winograd_2015}
Andrew Lavin and Scott Gray.
\newblock Fast algorithms for convolutional neural networks.
\newblock {\em 2016 IEEE Conference on Computer Vision and Pattern Recognition
  (CVPR)}, pages 4013--4021, 2016.

\bibitem{persephonee_2021}
Ilias Leontiadis, Stefanos Laskaridis, Stylianos~I. Venieris, and Nicholas~D.
  Lane.
\newblock It's always personal: Using early exits for efficient on-device cnn
  personalisation.
\newblock In {\em Proceedings of the 22nd International Workshop on Mobile
  Computing Systems and Applications}, HotMobile '21, page 15–21, New York,
  NY, USA, 2021. Association for Computing Machinery.

\bibitem{nin_Lin_2013}
Min Lin, Q. Chen, and Shuicheng Yan.
\newblock Network in network.
\newblock {\em CoRR}, abs/1312.4400, 2014.

\bibitem{augur_Lu_2019}
Zongqing Lu, Swati Rallapalli, Kevin Chan, Shiliang Pu, and Thomas~La Porta.
\newblock Augur: Modeling the resource requirements of convnets on mobile
  devices.
\newblock {\em IEEE Transactions on Mobile Computing}, 20(2):352--365, 2021.

\bibitem{thinet_luo_2017}
Jian{-}Hao Luo, Jianxin Wu, and Weiyao Lin.
\newblock Thinet: {A} filter level pruning method for deep neural network
  compression.
\newblock In {\em {IEEE} International Conference on Computer Vision, {ICCV}
  2017, Venice, Italy, October 22-29, 2017}, pages 5068--5076. {IEEE} Computer
  Society, 2017.

\bibitem{fft}
Micha{\"e}l Mathieu, Mikael Henaff, and Y. LeCun.
\newblock Fast training of convolutional networks through ffts.
\newblock {\em CoRR}, abs/1312.5851, 2014.

\bibitem{taylor-fo_pruning_molchanov_2019}
Pavlo Molchanov, Arun Mallya, Stephen Tyree, Iuri Frosio, and Jan Kautz.
\newblock Importance estimation for neural network pruning.
\newblock In {\em {IEEE} Conference on Computer Vision and Pattern Recognition,
  {CVPR} 2019, Long Beach, CA, USA, June 16-20, 2019}, pages 11264--11272.
  Computer Vision Foundation / {IEEE}, 2019.

\bibitem{dec_tree_ref}
J.~Ross Quinlan.
\newblock {\em C4.5: programs for machine learning}.
\newblock Morgan Kaufmann Publishers Inc., San Francisco, CA, USA, 1993.

\bibitem{dapr_rajagopal2020}
Aditya Rajagopal and Christos-Savvas Bouganis.
\newblock Now that i can see, i can improve: Enabling data-driven finetuning of
  cnns on the edge.
\newblock In {\em Proceedings of the IEEE/CVF Conference on Computer Vision and
  Pattern Recognition (CVPR) Workshops}, June 2020.

\bibitem{backprop_rumelhart_1986}
David~E. Rumelhart, Geoffrey~E. Hinton, and Ronald~J. Williams.
\newblock Learning representations by back-propagating errors.
\newblock {\em Nature}, 323:533--536, 1986.

\bibitem{mobilenetv2_Sandler2018}
Mark Sandler, Andrew Howard, Menglong Zhu, Andrey Zhmoginov, and Liang-Chieh
  Chen.
\newblock Mobilenetv2: Inverted residuals and linear bottlenecks.
\newblock {\em Proceedings of the IEEE Computer Society Conference on Computer
  Vision and Pattern Recognition}, pages 4510--4520, 1 2018.

\bibitem{vgg_Simonyan_15}
Karen Simonyan and Andrew Zisserman.
\newblock Very deep convolutional networks for large-scale image recognition.
\newblock In Yoshua Bengio and Yann LeCun, editors, {\em 3rd International
  Conference on Learning Representations, {ICLR} 2015, San Diego, CA, USA, May
  7-9, 2015, Conference Track Proceedings}, 2015.

\bibitem{googlenet_Szegedy2015}
Christian Szegedy, Wei Liu, Yangqing Jia, Pierre Sermanet, Scott Reed, Dragomir
  Anguelov, Dumitru Erhan, Vincent Vanhoucke, and Andrew Rabinovich.
\newblock Going deeper with convolutions.
\newblock In {\em Computer Vision and Pattern Recognition (CVPR)}, 2015.

\bibitem{mnasnet_Tan2018}
Mingxing Tan, Bo Chen, Ruoming Pang, Vijay Vasudevan, Mark Sandler, Andrew
  Howard, and Quoc~V. Le.
\newblock Mnasnet: Platform-aware neural architecture search for mobile.
\newblock {\em Proceedings of the IEEE Computer Society Conference on Computer
  Vision and Pattern Recognition}, 2019-June:2815--2823, 7 2018.

\bibitem{winograd1980arithmetic}
S. Winograd.
\newblock {\em Arithmetic Complexity of Computations}.
\newblock CBMS-NSF Regional Conference Series in Applied Mathematics. Society
  for Industrial and Applied Mathematics, 1980.

\bibitem{edge_survey_zhou_2019}
Z. Zhou, X. Chen, En Li, Liekang Zeng, Ke Luo, and Junshan Zhang.
\newblock Edge intelligence: Paving the last mile of artificial intelligence
  with edge computing.
\newblock {\em Proceedings of the IEEE}, 107:1738--1762, 2019.

\end{thebibliography}
}

\clearpage
\appendix
\newpage

\section{Profiling in PyTorch}
    \label{app:profiling-variables}
    This section details the practical implementation of data collection for the attributes described in Sec.\ref{sec:problem_definition}. 
    Code to support the description here is open-sourced in the tool associated with this work.
    PyTorch forward and backward hooks are attached to the network's convolution layers and at each hook the following values are recorded for pruning levels ${5x | x \in [0,18]}$ and batch sizes \{2,4,8,16,32,64,70,80,90,100,110,120,128,140,150,160,170,\\180,190,200,210,220,230,240,256\}.
    
    Results provided in the paper were for both the NVIDIA Jetson TX2 and the NVIDIA RTX 2080Ti.
    $\Phi$ is profiled in the same manner for the two systems (embedded and server):
    %
    
    \paragraph{Mini-batch training latency ($\Phi$)} is profiled using the \texttt{torch.cuda.Events} API as this allows to time asynchronous GPU events.\newline
    
    \par The total training memory attribute is profiled differently for embedded (NVIDIA Jetson TX2) and server (NVIDIA RTX 2080Ti) systems due to the difference in memory management where the embedded device has 
    a unified memory space, where the server system has an independent memory space for the GPU. 
    
    \paragraph{Memory Used ($\Gamma$)}
        \begin{itemize}
            \item \textbf{NVIDIA Jetson TX2} - Profiled using information from the \texttt{/proc/meminfo} system file.
            As the Jetson has a unified memory, information in this file accounts for both GPU and CPU memory.
            \begin{equation*}
                    \begin{split}
                        &TotalMemUsed = \\
                        &MemTotal - MemFree - Buffers - Cached
                    \end{split}
                \end{equation*}
            The terms on the RHS of the equation correspond directly to values found in \texttt{/proc/meminfo} file.
            
            \item \textbf{NVIDIA RTX 2080Ti} - Profiled using the \texttt{pynvml} tool with the command \texttt{nvmlDeviceGetMemoryInfo.used}.
        \end{itemize} 
    The system records the maximum value of $\Gamma$ observed until the point of profiling.
    
    %

\newpage

\section{Model Construction}
    \label{app:model_terms}
    \begin{figure*}[t]
    \begin{minipage}[b]{\linewidth}
    \resizebox{\textwidth}{!}{
        \begin{tabular}{cccc}
            \subfloat[ResNet18 $\Gamma$]{\includegraphics[width=0.3\textwidth, bb=11 9 416 305]{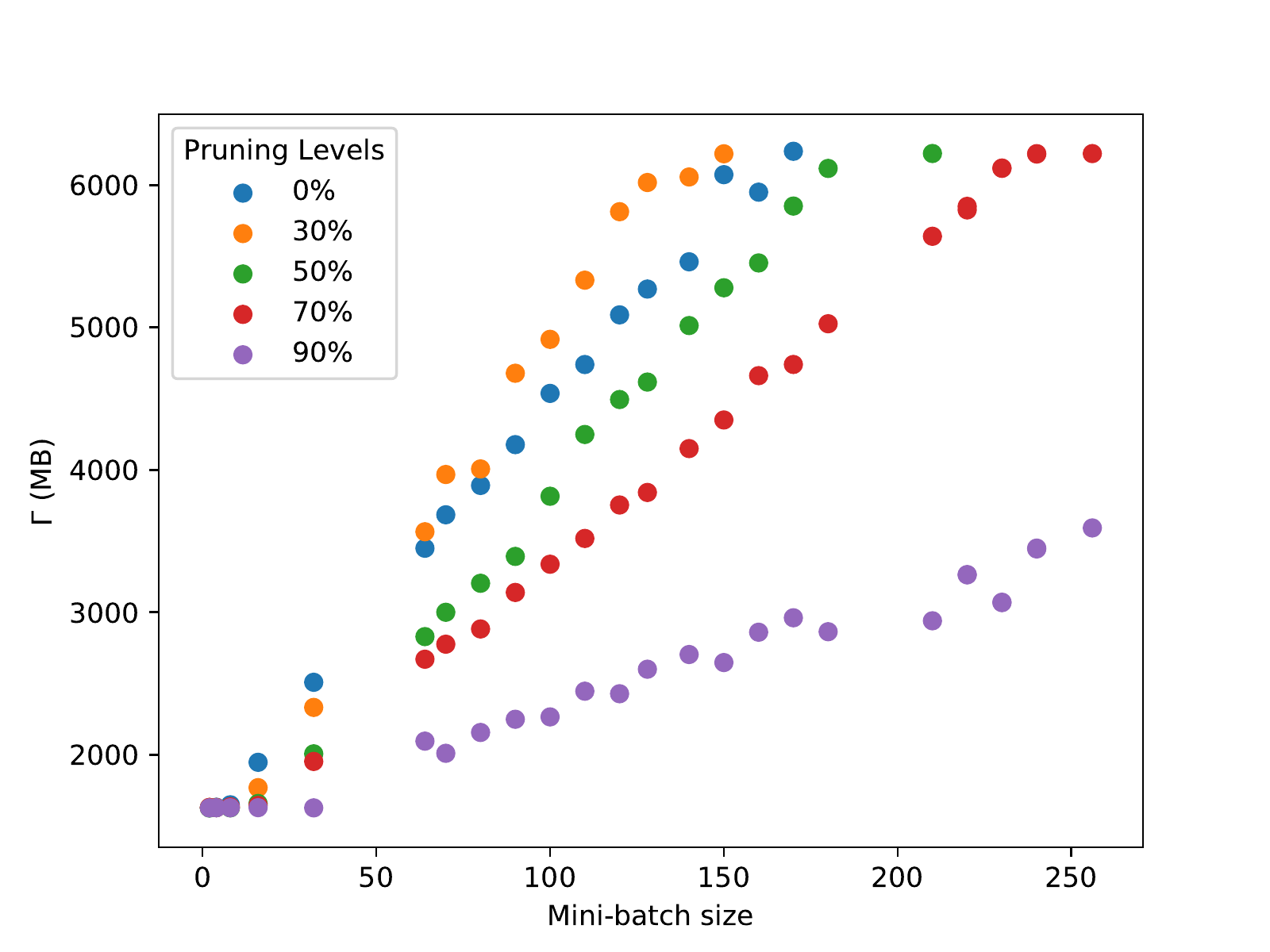} \label{fig:resnet18_mem_used}} &
            \subfloat[MobileNetV2 $\Gamma$]{\includegraphics[width=0.3\textwidth, bb=11 9 416 305]{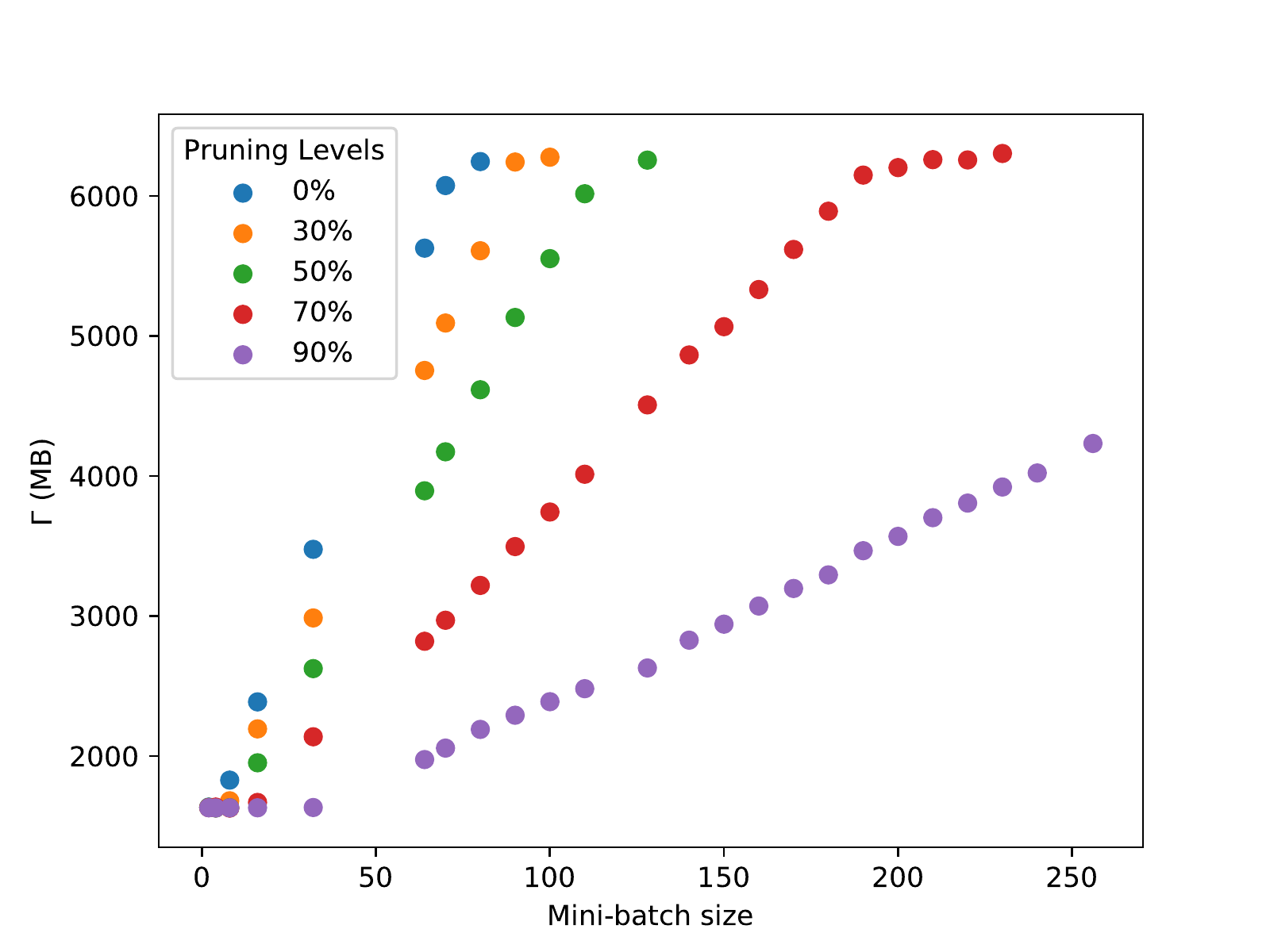} \label{fig:mobilenetv2_mem_used}} &
            \subfloat[SqueezeNet $\Phi$]{\includegraphics[width=0.3\textwidth, bb=11 9 416 305]{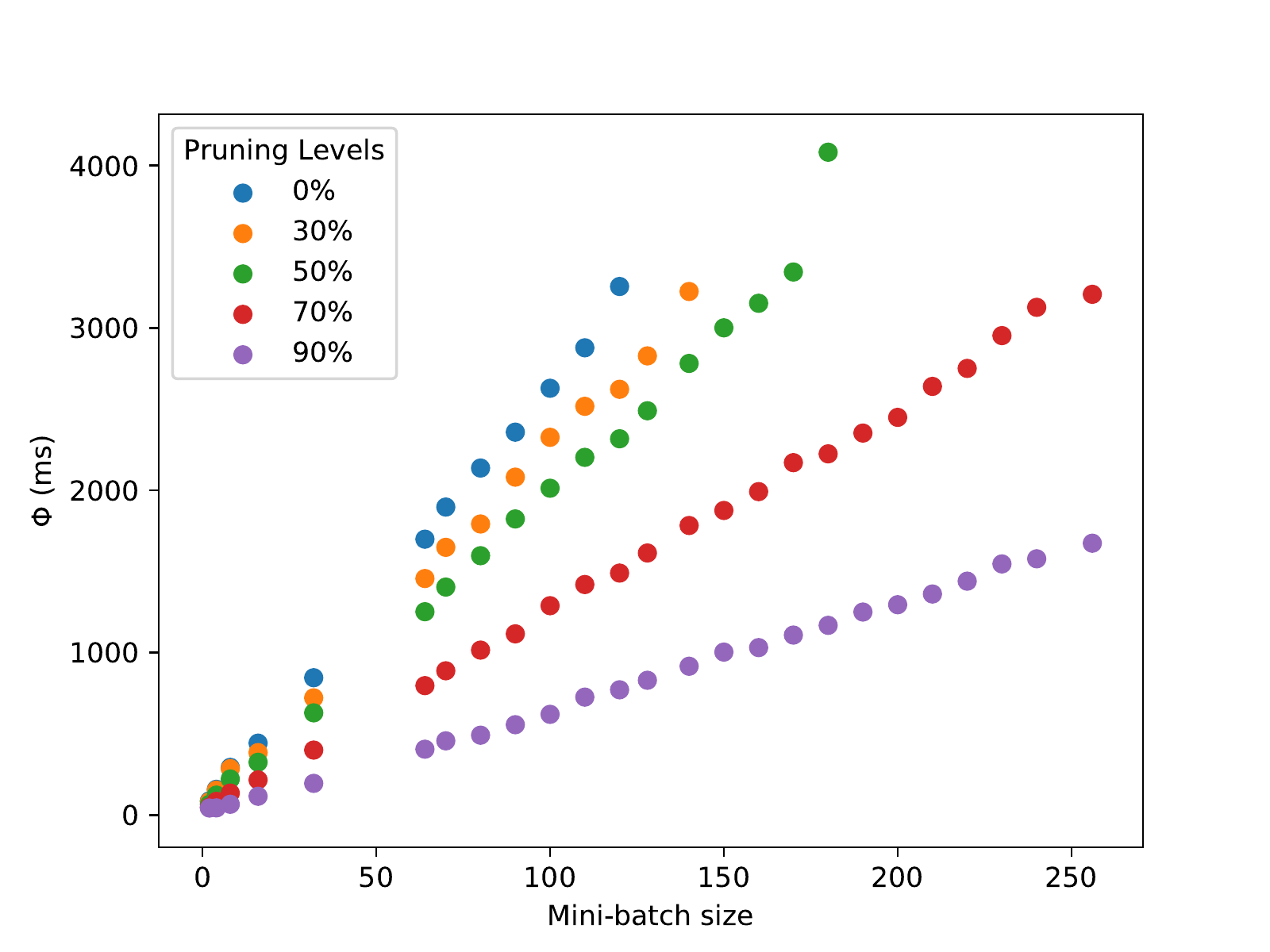} \label{fig:squeezenet_latency}} &
            \subfloat[MnasNet $\Phi$]{\includegraphics[width=0.3\textwidth, bb=11 9 416 305]{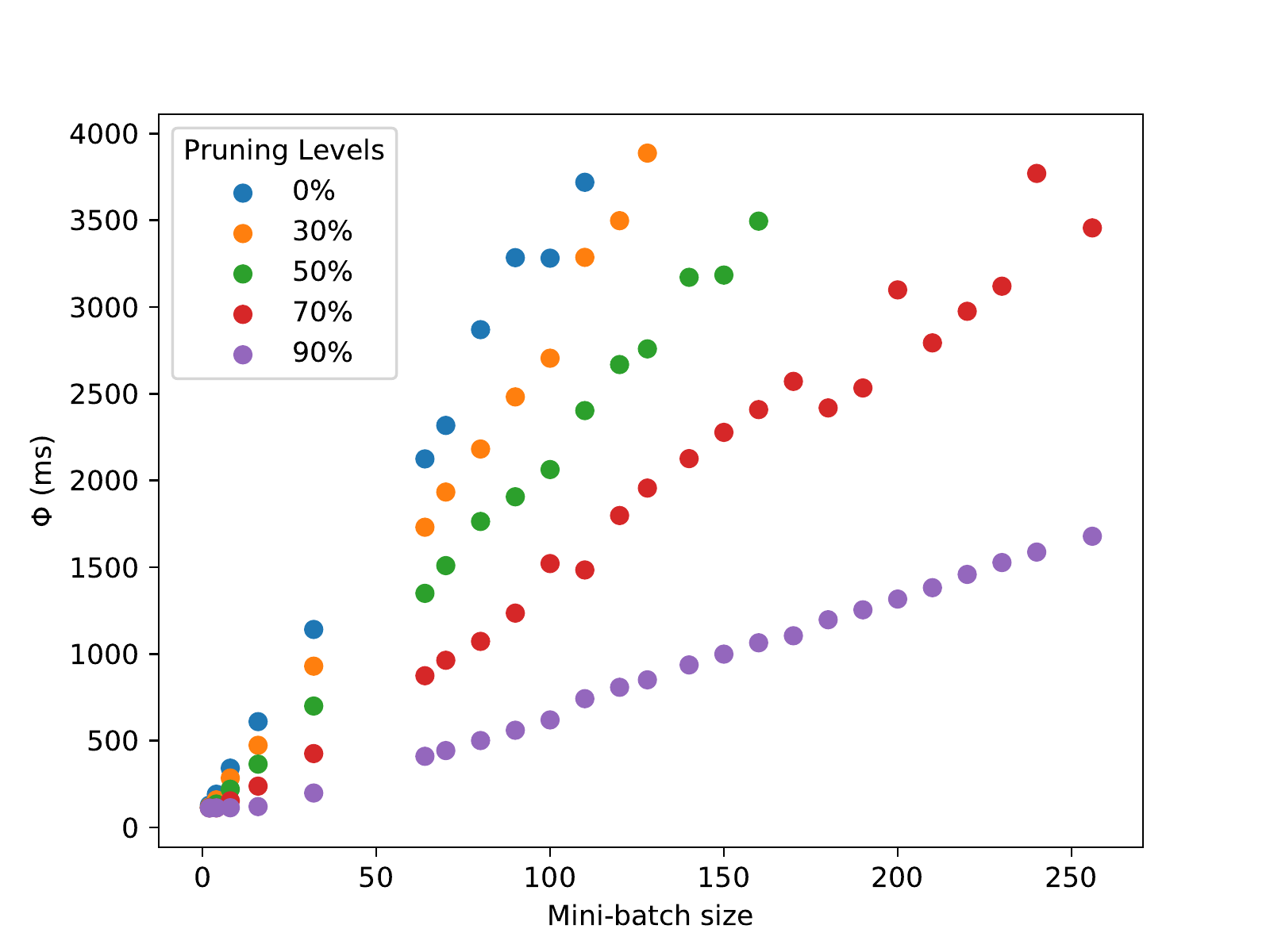} \label{fig:mnasnet_latency}} 
        \end{tabular}
    }
    \caption{$\Gamma$ and $\Phi$ for 4 different networks that were profiled when training 3x224x224 sized inputs on the NVIDIA Jetson TX2. x-axis is mini-batch size and y-axis is the attribute value.}
    \label{fig:profiled_data}
    \end{minipage}
\end{figure*}

    \subsection{Initial Profiling}
        Graphs corresponding to the profiling described in Sec.\ref{sec:perf_model} for ResNet18, MobileNetV2, SqueezeNet and MnasNet for the modelled attributes are shown in Fig.\ref{fig:profiled_data}.
        As discussed, they display linearity with batch size and varying linear fit dependent on the pruning level. 
     
    \subsection{Complete list of model features}
        This section lists the features described in Sec.\ref{sec:model_terms} along with various summations of these features that are used to develop the performance model.
        
        Consider a CNN where each convolutional layer $l \in \mathcal{L}$ has $n_l$ filters of size $m_l \times k_l \times k_l$. 
        Let layer $l$ have stride $s_l$, padding $p_l$ and groups $g_l$.
        Let the IFM to this layer have dimensions $bs \times m_l \times ip_l \times ip_l$, the weights $n_l \times \frac{m_l}{g_l} \times k_l \times k_l$ and the OFM $bs \times n_l \times op_l \times op_l$ where $bs$ is the batch size of training.
        The OFM spatial dimensions $op_l$ can be calculated from other variables using the equation $op_l = 1 + \lfloor \frac{ip_l + 2p_l - k_l}{s_l} \rfloor$.
        
        \subsubsection{Tensor Allocations}
            The features in this section describe the sizes of tensors that constitute the IFM, weights and OFM; and each of their gradients on a per-layer basis. 
            \begin{enumerate}
                \item   $mem_{w} = n_l \cdot \frac{m_l}{g_l} \cdot k_l^2$ 
                \item   $mem_{w_{grad}} = bs \cdot n_l \cdot \frac{m_l}{g_l} \cdot k_l^2$
                \item   $mem_{ifm_{grad}} = mem_{ifm} = bs \cdot m_l \cdot ip_l^2$ 
                \item   $mem_{ofm_{grad}} = mem_{ofm} = bs \cdot n_l \cdot op_l^2$
                \item   $mem_{w} + mem_{w_{grad}} + mem_{ifm_{grad}} + mem_{ofm_{grad}}$ 
            \end{enumerate}
        
        The following sections model the memory consumption and operations for the Matrix Multiplication, FFT and Winograd based convolution algorithms on a per layer basis. 
        Each feature models one of Eq.\ref{eq:forward_pass} ($fwd$), the forward pass; Eq.\ref{eq:backward_x} ($bwd_x$), the computation of the gradient w.r.t input; or Eq.\ref{eq:backward_w}, the computation of the gradient w.r.t weights.
        
        \subsubsection{Matrix Multiplication based convolution}
            The features in this section are obtained by modelling the sizes of the matrices that are used to perform a convolution through matrix multiplications, and calculating the memory and operations required to store these matrices and perform the multiplication. 
            The $total$ features correspond to the MATMUL strategy where the entire $im2col$ matrix is stored in memory, while $idx$ features correspond to the strategy of storing only the necessary indices as described in Sec.\ref{sec:matmul_terms}.
            
            As an example, the forward pass (Eq.\ref{eq:forward_pass}) involves a multiplication between the $im2col$ IFM matrix of size $((bs \times op_l^2) \times (k_l^2 \times m_l))$ and a reshaped weights matrix of size $((k_l^2 \times m_l) \times (n_l))$.
            Using $k_l^2 \times m_l$ as the common dimension and calculating the memory and operations for a matrix multiplication gives the features $mem\_i2c_{fwd^{total}}^{mm}$ and $ops_{bwd_x}^{mm}$.
            As discussed in Sec.\ref{sec:matmul_terms}, the index of each of the $op_l^2$ output pixels needs to be stored thus giving the $mem\_i2c_{fwd^{index}}^{mm}$ feature. 
            Similar arguments applied to Eq.\ref{eq:backward_x} and \ref{eq:backward_w} give the remaining features in this section.
            \begin{enumerate}[resume]
                \item   $mem\_i2c_{fwd^{total}}^{mm} = bs \cdot op_l^2 \cdot k_l^2 \cdot m_l$
                \item   $mem\_i2c_{bwd_w^{total}}^{mm} = bs \cdot op_l^2 \cdot k_l^2 \cdot \frac{m_l}{g_l}$
                \item   $mem\_i2c_{fwd^{index}}^{mm} = i2c_{bwd_w^{index}}^{mm} = bs \cdot op_l^2$
                \item   $mem\_i2c_{bwd_x^{total}}^{mm}  = bs \cdot ip_l^2 \cdot k_l^2 \cdot m_l$
                \item   $mem\_i2c_{bwd_x^{index}}^{mm}  = bs \cdot ip_l^2$
                \item   $mem\_i2c_{fwd^{total}}^{mm} + mem\_i2c_{bwd_w^{total}}^{mm} + mem\_i2c_{bwd_x^{total}}^{mm}$
                \item   $2 \times mem\_i2c_{fwd^{index}}^{mm} + mem\_i2c_{bwd_x^{index}}^{mm}$
                
                \item   $ops_{fwd}^{mm} = ops_{bwd_w}^{mm} = bs \cdot n_l \cdot op_l^2 \cdot k_l^2 \cdot \frac{m_l}{g_l}$
                \item   $ops_{bwd_x}^{mm} = bs \cdot m_l \cdot ip_l^2 \cdot k_l^2 \cdot n_l$
                \item   $2 \times ops_{fwd}^{mm} + ops_{bwd_x}^{mm}$
            \end{enumerate}
        
        \subsubsection{FFT based convolution}
            The features in the section, apart from the summations, have been obtained from \cite{fft} where there is a detailed breakdown of the memory consumption and operations of using the FFT algorithm for matrix multiplication. 
            \begin{enumerate}[resume]
                \item   $mem\_w_{fwd}^{fft} = n_l \cdot \frac{m_l}{g_l} \cdot ip_l \cdot (1 + ip_l)$
                \item   $mem\_ifm_{fwd}^{fft} = ifm_{bwd_w}^{fft} = bs \cdot m_l \cdot ip_l \cdot (1+ip_l)$
                \item   $mem\_ofm_{bwd_w}^{fft} = bs \cdot n_l \cdot ip_l \cdot (1+ip_l)$
                \item   $mem\_w_{bwd_x}^{fft} = n_l \cdot \frac{m_l}{g_l} \cdot op_l \cdot (1+op_l)$
                \item   $mem\_ofm_{bwd_x}^{fft} = bs \cdot n_l \cdot op_l \cdot (1+op_l)$
                \item   $mem\_w_{fwd}^{fft} + mem\_ifm_{fwd}^{fft}$ \label{fft:sum1}
                \item   $mem\_ofm_{bwd_x}^{fft} + mem\_ofm_{bwd_w}^{fft}$ \label{fft:sum2}
                \item   $mem\_ofm_{bwd_w}^{fft} + mem\_ifm_{fwd}^{fft}$ \label{fft:sum3}
                \item   $\ref{fft:sum1} + \ref{fft:sum2} + \ref{fft:sum3}$
                        
                \item   $ops_{fwd}^{fft} = ip_l^2 \cdot log(ip_l) \cdot (bs \cdot (m_l + n_l) + n_l \cdot \frac{m_l}{g_l}) \\ + bs \cdot n_l \cdot m_l \cdot ip_l^2$
                \item   $ops_{bwd_{x}}^{fft} = op_l^2 \cdot log(op_l) \cdot (bs \cdot (m_l + n_l) + n_l \cdot \frac{m_l}{g_l}) \\ + bs \cdot n_l \cdot m_l \cdot op_l^2$
                \item   $ops_{bwd_{w}}^{fft} = ip_l \cdot log(ip_l^2) \cdot (bs \cdot (m_l + n_l) + n_l \cdot \frac{m_l}{g_l}) \\ + bs \cdot n_l \cdot m_l \cdot ip_l^2$
                \item   $ops_{fwd}^{fft} + ops_{bwd_{x}}^{fft} + ops_{bwd_{w}}^{fft}$
            \end{enumerate}
        
        \subsubsection{Winograd convolution}
            Each of the features described in this section model Eq.\ref{eq:winograd} for each of Eq.\ref{eq:forward_pass},\ref{eq:backward_x},\ref{eq:backward_w}. 
            Consider the case of Eq.\ref{eq:backward_x}. 
            $\frac{\delta L}{\delta y}$ is a matrix of size $(bs \times n_l \times op_l \times op_l)$ and $w_{nm}$ is a matrix of size $(n_l \times m_l \times k_l \times k_l)$.
            The term $d$ in Eq.\ref{eq:winograd}, is one of the $\lceil \frac{op_l}{q} \rceil^2$ tiles of $\frac{\delta L}{\delta y}$ and $g$ is one of the $\lceil \frac{k_l}{r} \rceil^2$ tiles of $w$.
            Assuming parallelism over $bs$, $m_l$ and $\lceil \frac{op_l}{q} \rceil^2$ gives the feature $mem_{bwd_x}^{wino}$ and accounting for $n_l \cdot \lceil \frac{k_l}{r} \rceil^2$ accumulations gives the feature $ops_{fwd}^{wino}$.
            Applying similar arguments to Eq.\ref{eq:forward_pass} and \ref{eq:backward_w} gives the remainder of the features. 
            
            The following features are applied twice for $(q \times r)$ of $(4 \times 3)$ and $(3 \times 2)$ which both profiling and \cite{cudnn_review_2019} showed to be the most commonly used sizes of winograd convolutions by CuDNN· 
            \begin{enumerate}[resume]
                \item   $mem_{fwd}^{wino} = bs \cdot n_l \cdot \lceil \frac{ip_l}{q} \rceil^2 \cdot 3 \cdot (q+r-1)^2$
                \item   $mem_{bwd_x}^{wino} = bs \cdot m_l \cdot \lceil \frac{op_l}{q} \rceil^2 \cdot 3 \cdot (q+r-1)^2$
                \item   $mem_{bwd_w}^{wino} = bs \cdot n_l \cdot \frac{m_l}{g_l} \cdot \lceil \frac{ip_l}{q} \rceil^2 \cdot 3 \cdot (q+r-1)^2$
                \item   $mem_{fwd}^{wino} + mem_{bwd_x}^{wino}$ \label{wino:sum1}
                \item   $mem_{fwd}^{wino} + mem_{bwd_w}^{wino}$ \label{wino:sum2}
                \item   $mem_{bwd_w}^{wino} + mem_{bwd_x}^{wino}$ \label{wino:sum3}
                \item   $\ref{wino:sum1} + \ref{wino:sum2} + \ref{wino:sum3}$
                
                \item   $ops_{fwd}^{wino} = bs \cdot n_l \cdot \frac{m_l}{g_l} \cdot \lceil \frac{ip_l}{q} \rceil^2 \cdot \lceil \frac{k}{r} \rceil^2 \cdot (q+r-1)^2$
                \item   $ops_{bwd_x}^{wino} = bs \cdot m_l \cdot n_l \cdot \lceil \frac{op_l}{q} \rceil^2 \cdot \lceil \frac{k}{r} \rceil^2 \cdot (q+r-1)^2$
                \item   $ops_{bwd_w}^{wino} = bs \cdot n_l \cdot \frac{m_l}{g_l} \cdot \frac{m_l}{g_l} \cdot \lceil \frac{ip_l}{q} \rceil^2 \cdot \lceil \frac{op_l}{r} \rceil^2 \cdot (q+r-1)^2$
                \item   $ops_{fwd}^{wino} + ops_{bwd_x}^{wino}$ \label{wino_ops:sum1}
                \item   $ops_{fwd}^{wino} + ops_{bwd_w}^{wino}$ \label{wino_ops:sum2}
                \item   $ops_{bwd_x}^{wino} + ops_{bwd_w}^{wino}$ \label{wino_ops:sum3}
                \item   $\ref{wino_ops:sum1} + \ref{wino_ops:sum2} + \ref{wino_ops:sum3}$
            \end{enumerate}

\newpage

\section{Training on a "basis" of networks}
    \label{app:basis_investigation}
    \begin{table}[t]
    \vskip 0.15in
    \begin{center}
    \begin{small}
    \begin{sc}
    \resizebox{0.5\textwidth}{!}{
        \begin{tabular}{ll}
            \toprule
            \multicolumn{1}{c}{\textbf{Building Block}} &
            \multicolumn{1}{c}{\textbf{Networks}} \\
            \midrule
            Residual                &   ResNet18, ResNet50 \\
            Depth-wise Separable    &   MobileNetV2, MnasNet \\
            Fire                    &   SqueezeNet \\
            Inception               &   GoogLeNet \\  
            \bottomrule
        \end{tabular}
    }
    \end{sc}
    \end{small}
    \end{center}
    \vskip -0.1in
    \caption{Summary of architectural building blocks and networks that utilise them}
    \label{tab:building_blocks}
\end{table}

    This section provides additional results for Sec.\ref{sec:eval_unseen_net_unseen_pl} which investigated if training could be performed on data from a representative "basis" of networks and predict attributes on other 
    networks not present in the basis but with similar building blocks to those in the basis.
    The following discussion illustrates the similarity of building blocks between networks in the basis and those not present in it.
    
    Tab.\ref{tab:building_blocks} details various commonly used building blocks and the networks that utilise them. 
    Both the Fire and Inception modules employ a "branch-and-concatenate" computation structure and have $1 \times 1$ and $3 \times 3$ convolutions. 
    The Fire module has 2 branches, whereas the Inception module has 4 branches and a $5 \times 5$ convolution.
    ResNet18 is made of "Basic-block residuals" which have two $3 \times 3$ convolutions, whereas the deeper ResNet50 is made of "Bottleneck residuals" which have 3 convolutions (one $3 \times 3$ and two $1 \times 1$). 
    MobileNetV2 and MnasNet are made up of the same depth-wise separable inverted residual module, but have different depths. 
    Thus architecture pairs from most to least similar are: MobileNetV2 and MnasNet; ResNet18 and ResNet50; and SqueezeNet and GoogLeNet.
    The results of this investigation are inline with this observation where MnasNet performed the best followed by ResNet50 and then GoogLeNet. 

\newpage


\section{On-device OFA Case Study Subsets}
    \label{app:subset_details}
    This section provides details on the classes present in each of the 4 subsets used in the on-device OFA case study presented in Sec.\ref{sec:case_study}. 
    The subsets were extracted from sub classes of the ImageNet dataset and were created to emulate objects, people, buildings and animals that might observed in different environments that an autonomous vehicle could encounter.
    The details of the classes present in each subset are as follows:
    \begin{itemize}
        \item \textbf{City} (185 classes) : ambulance, trash can, station wagon, tandem bike, taxi, car mirror, car wheel, convertible, crane, electric locomotive, fire engine, fire truck, garbage truck, petrol pump,
                radiator grille, jeep landrover, rickshaw, lawn mower, limousine, mailbox, manhole cover, minibus, minivan, Model T, moped, scooter, mountain bike, moving van, parking meter, passenger car / coach, 
                pay-phone, pickup truck, police car, race car, school bus, shopping cart, snowplow, sports car, steel arch bridge, tram, suspension bridge, tow truck, trolley bus, street sign, traffic light, palace, 
                mosque, church building, castle, boathouse, tirumphal arch, academic gown/robe, cardigan, fur coat, gown, jersey / t-shirt, suit, sunglasses, sweatshirt, trench coat, umbrella, swan, dogs, red fox, cats

        \item \textbf{Motorway} (26 classes) : station wagon, bullet train, car mirror, car wheel, convertible, electric locomotive, petrol pump, jeep landrover, minibus, minivan, mobile home, Model T, moving van, 
                passenger car / coach, pay-phone, pickup truck, police car, race car, recreational vehicle, snowplow, sports car, tow truck, trailer truck, street sign, water tower

        \item \textbf{Country-side} (204 classes) : wheelbarrow, station wagon, bullet train, taxi, car mirror, car wheel, convertible, electric locomotive, freight car, garbage truck, petrol pump, radiator grille, 
                half track, horse cart, jeep landrover, lawn mower, mailbox, manhole cover, minibus, minivan, mobile home, Model T, moped, scooter, mountain bike, moving van, oxcart, parking meter, passenger car / coach, 
                pay-phone, picket-fence, pickup truck, plough, police car, race car, recreational vehicle, school bus, snowmobile, snowplow, sports car, steel arch bridge, tank, thatched roof, tile roof, tow truck, tractor, 
                trailer truck, worm fence, street sign, traffic light, hay, palace, mosque, church building, castle, lighthouse, barn, viaduct, water tower, cardigan, fur coat, gown, sarong, jersey / t-shirt, suit, sunglasses, 
                sweatshirt, swimming trunks, trench coat, umbrella, cock, hen, quail, goose, swan, dogs, red fox, cats, rabbits, ram, sheep

        \item \textbf{Off-road} (26 classes) : mobile home, mountain bike, oxcart, pickup truck, plough, snowmobile, tank, tractor, hay, ostrich, iguana, alligator, wallaby, koala, wombat, brown bear, black bear, hog, 
                wild boar, ox, water buffalo, bison, wild deer 
    \end{itemize}
    There are fewer written categories than the number of classes stated as some categories such as "dogs" have many ILSVRC'12 classes within them.
    There are also overlap of classes between the subsets as would be expected.

\end{document}